\documentclass{new_tlp}

\usepackage[semibold,scale=0.875]{sourcecodepro}
\usepackage[utf8]{inputenc}
\usepackage{amsmath}
\usepackage{amssymb}
\usepackage{url}

\def\SPos{{\rm SPos}}

\def\ar{\leftarrow}
\def\rar{\rightarrow}
\def\lrar{\leftrightarrow}
\def\no{\emph{not}}

\def\beq{\begin{equation}}
\def\eeq#1{\label{#1}\end{equation}}
\def\ba{\begin{array}}
\def\ea{\end{array}}

\def\p2f{\hbox{p2f}}

\newtheorem{theorem}{Theorem}
\newtheorem{lemma}{Lemma}

\title{Positive Dependency Graphs Revisited}

\author[Jorge Fandinno and Vladimir Lifschitz]{%
   Jorge Fandinno$^1$ and Vladimir Lifschitz$^2$\\
   $^1$University of Nebraska Omaha, USA\\
   \email{jfandinno@unomaha.edu}\\
   University of Texas at Austin, USA\\
   \email{lifschitzv@gmail.com}
}

\begin{document}
\maketitle



\begin{abstract}
Theory of stable models is the
mathematical basis of answer set programming.
Several results in that theory refer to the
concept of the positive dependency graph of a logic program.  We
describe a modification of that concept and show that
the new understanding of positive dependency makes it possible to strengthen some of these
results.
\end{abstract}

\section{Introduction}

This note contributes to the theory of stable models, which serves as the
mathematical basis of answer set programming
\cite{mar99,nie99,lif19a}.
Several results in that theory refer
to ``positive dependencies'' between atoms in a logic program---the
idea used by
Fran\c{c}ois Fages \cite{fag94} for the purpose of describing the
relationship between program completion~\cite{cla78} and stable
models~\cite{gel88}.  It was applied later to the study of loops and
to designing the answer set solver {\sc assat} \cite{lin04}, and found other
applications.

For a program consisting of rules of the form
\beq
H\ar B_1,\dots,B_m,\no\ B_{m+1},\dots,\no\ B_n,
\eeq{crule}
where $H,B_1,\dots,B_n$ are propositional atoms, the \emph{positive dependency
graph} is defined as the directed graph such that
\begin{itemize}
\item
 its vertices are the atoms occurring in the program, and
\item
 its edges go from~$H$ to $B_1,\dots,B_m$ for all rules~(\ref{crule})
 of the program.
\end{itemize}
For example, the positive dependency graph of the program
\beq
\ba l
q\ar p,\\
p\ar q,\no\ r
\ea\eeq{prog1}
has two edges, $(q,p)$ and $(p,q)$.

In the early days of answer set programming, the syntactic form of every
rule of a program was similar to~(\ref{crule}), so that the definition of
the positive dependency graph above was applicable to all grounded
programs.  Later on, the syntax of rules was extended in several ways.
In one of these generalizations, reviewed in Section~\ref{sec:ssc} below,
rules are replaced by arbitrary propositional formulas~\cite{fer05}.
Rule~(\ref{crule}) can be viewed as a special case---as alternative
notation for the implication
$$
B_1\land\cdots\land B_m\land\neg B_{m+1}\land\cdots\neg B_n \rar H.
$$
This degree of generality is important in connection with the use of
aggregates, such as the cardinality of a set, in the body of a rule
\cite[Section~4]{fer05}.

A generalization of the definition of the positive dependency graph to
propositional formulas~\cite{fer06} and further generalizations have
been used for several purposes:
\begin{enumerate}
\item[(i)]
 to extend Fages' theorem on tight programs \cite{fag94} to
 first-order formulas \cite{fer09} and to infinitary propositional
 formulas~\cite{lif13a},
\item[(ii)]
 to extend the theory of loops~\cite{lin04}
 to arbitrary propositional formulas \cite{fer06},
\item[(iii)]
 to investigate a logic programming counterpart of pointwise
 circumscription~\cite{lif86} in the context of first-order
 formulas~\cite{fer09},
\item[(iv)]
 to extend the process of symmetric splitting \cite{oik08}
 to first-order formulas \cite{fer09a} and to infinitary propositional
 formulas \cite{har16a}.
\end{enumerate}

In this note, we reexamine the definition of the positive dependency graph
used in these publications and argue that a different interpretation of
positive dependency would be more appropriate in two of these
research lines---in
those listed above under~(i) and~(iii).
Two theorems on properties of modified dependency graphs are stated in Sections 4.1 and 4.3 and proved in Section 5.  The possibility of extending Fages' theorem along the lines of Theorem 1 is used in the proof of a theorem on the verification of locally tight programs~\cite{fanlif21a}.

\section{Review: Stable Models of Propositional Theories} \label{sec:ssc}

We assume that formulas are built from propositional atoms
and the symbol~$\bot$ using the binary connectives $\land$,
$\lor$, $\rar$; $\neg F$ stands for $F\rar\bot$, and $F\lrar G$ stands for
$(F\rar G)\land (G\rar F)$.  A \emph{propositional theory} is a set of
formulas.  An \emph{interpretation} is a set of atoms; we
identify an interpretation~$I$ with the truth assignment that maps the
elements of~$I$ to \emph{true} and all other atoms to \emph{false}.

The \emph{reduct} $F^I$ of a formula~$F$ with respect to an interpretation~$I$
is the formula obtained from~$F$ by replacing every maximal subformula of~$F$
that is not satisfied by~$I$ with~$\bot$ \cite[Section~2.1]{fer05}.  The
reduct~$T^I$ of a propositional theory~$T$ is the set of the
reducts~$F^I$ of all formulas~$F$ in~$T$.
An interpretation~$I$ is a \emph{stable model} of a propositional
theory~$T$ if it is minimal (with respect to set inclusion) among
the models of~$T^I$.

Consider, for instance, the formulas
\beq
\ba l
p\rar q,\\
q\land\neg r \rar p,
\ea\eeq{p1}
corresponding to rules~(\ref{prog1}).  The reduct of each of them
with respect to the interpretation~$\emptyset$ is the tautology $\bot\rar\bot$;
since~$\emptyset$ is a minimal model of this tautology, it is a stable model
of theory~(\ref{p1}).  The reduct of~(\ref{p1}) with respect to $\{p,q\}$
is
$$\ba l
p\rar q,\\
q\land\neg \bot \rar p.
\ea$$
The interpretation $\{p,q\}$ is a model of this reduct, but it is not minimal:
its subset~$\emptyset$ is a model of the reduct as
well.  Consequently $\{p,q\}$ is not a stable model of~(\ref{p1}).

It is easy to check by induction that an interpretation~$I$ satisfies the
reduct~$F^I$ if and only if it satisfies~$F$.  It follows that every stable
model of a propositional theory~$T$ is a model of~$T$.

It is clear also that every atom occurring in~$F^I$ belongs to~$I$.

\section{A Tale of Two Graphs} \label{sec:tale}

A \emph{nondisjunctive rule} is an implication whose consequent is an atom.
Take a set~$T$ of nondisjunctive rules.  What graph will we designate as
the positive dependency graph of~$T$?  As far as the set of vertices is
concerned, the decision is straightforward---we will include all atoms that
occur in the members
\beq
\emph{Body}\rar H
\eeq{rule}
of~$T$.  How will we choose the edges of the graph?  For every
formula~(\ref{rule}) in~$T$, the graph will include edges going
from~$H$ to some of the atoms occurring in \emph{Body}.
But how will we decide which of the atoms
occurring in~\emph{Body} to choose as the heads of edges?

A subformula of a formula~$F$ is called \emph{strictly positive} if it does not
belong to the antecedent of any implication.  For instance, in a conjunction
of literals
$$B_1\land\cdots\land B_m\land\neg B_{m+1}\land\cdots\neg B_n$$
the atoms $B_1,\dots,B_m$ are strictly positive, and the atoms
$B_{m+1},\dots,B_n$ are not (recall that $\neg B_i$ is shorthand for the
implication $B_i\rar\bot$).
In our more general definition of the positive dependency graph
it would be natural to include, for every member~(\ref{rule}) of~$T$,
the edges from~$H$ to all atoms that have
\begin{center}
at least one strictly positive occurrence in \emph{Body}.
\end{center}
We will denote the graph formed from~$T$ according to this rule by
$G^{sp}(T)$.  (The superscript
\emph{sp} stands for \emph{strictly positive.})

However, the publications mentioned in the introduction
\cite{fer06,fer09,lif13a,fer09a,har16a} use a different, and more
complicated, definition of the positive dependency graph.
A subformula of a formula~$F$ is called
\begin{itemize}
 \item
   \emph{positive} if the number of implications containing it in the antecedent
   is even, and
 \item
   \emph{nonnegated} if it does not belong to the antecedent of any
   implication with the consequent~$\bot$.
\end{itemize}    
The graph designated as the positive
dependency graph of~$T$ in the publications mentioned above has
the same vertices as $G^{sp}(T)$, but its edges go from~$H$ to
all atoms that have
\begin{center}
at least one positive nonnegated occurrence in \emph{Body}
 \end{center}
for all members~(\ref{rule}) of~$T$.  We will denote this graph by
$G^{pnn}(T)$.  (The superscript \emph{pnn} stands for \emph{positive
  nonnegated}.)
 It is clear that $G^{sp}(T)$ is a subgraph of $G^{pnn}(T)$.
For example, if~$T$ is
$$((p\rar q)\rar r)\rar s$$
then the only edge of $G^{sp}(T)$ is $(s,r)$; $G^{pnn}(T)$ has two edges,
$(s,r)$ and $(s,p)$.

\section{Which Graph Is Right for Your Problem?}

The definitions of $G^{sp}$ and $G^{pnn}$ in Section~\ref{sec:tale}
are limited to sets of nondisjunctive rules.  We will now extend them to
arbitrary propositional theories; this generalization will be used in
Sections~\ref{ssec:l}--\ref{ssec:s}.

A strictly positive occurrence of an implication $\emph{Body}\rar\emph{Head}$
in a formula~$F$ is called a \emph{rule} of~$F$.
For any propositional theory~$T$, by~$G^{sp}(T)$ we denote the directed
graph such that \begin{enumerate}
 \item[(a)]
 its vertices are the atoms occurring in the members of~$T$, and
\item[(b)]
 for every rule $\emph{Body}\rar\emph{Head}$ of any member of~$T$,
 it includes the edge $(H,B)$ for every atom~$B$ that
 has at least one strictly positive occurrence in \emph{Body} and
 every atom~$H$ that has at least one strictly positive occurrence in
 \emph{Head}.
\end{enumerate}
By~$G^{pnn}(T)$ we denote the directed graph satisfying conditions~(a) and
\begin{enumerate}
\item[(b$'$)]
 for every rule $\emph{Body}\rar\emph{Head}$ of any member of~$T$,
 it includes the edge $(H,B)$ for every atom~$B$ that
 has at least one positive nonnegated occurrence in \emph{Body} and
 every atom~$H$ that has at least one strictly positive occurrence in
 \emph{Head}.
\end{enumerate}

For any formula~$F$, we will write $G^{sp}(\{F\})$
as~$G^{sp}(F)$, and similarly for~$G^{pnn}$.

\subsection{Supported Models}\label{ssec:sup}

A model~$I$ of a set~$T$ of nondisjunctive rules is \emph{supported} if
every atom~$A$ in~$I$ is the consequent of some member
$\emph{Body}\rar A$ of~$T$ such that~$I$ satisfies~$\emph{Body}$.
Supported models are important because of their relation to
program completion \cite{cla78,llo84a}: for any finite set~$T$ of
nondisjunctive rules, an interpretation~$I$ is a model of the completion
of~$T$ if and only if~$I$ is a supported model of~$T$ \cite{apt88}.

Every stable model of a set of nondisjunctive rules is supported, but the
converse is, generally, not true.  For instance, $\{p,q\}$ is a supported
model of~(\ref{p1}), but it is not stable.  From published work on
generalizations of Fages' theorem we know that the stability of all
supported models can be asserted for the sets~$T$ of nondisjunctive rules
such that the graph $G^{pnn}(T)$ has no infinite paths
\cite[Electronic
Appendix~B]{lif13a}. (For finite~$T$, this is the same as assuming
that the graph is acyclic.)  We will show that the graph $G^{sp}(T)$
has the same property:

\begin{theorem}\label{th1}
 For any set~$T$ of nondisjunctive rules, if the graph $G^{sp}(T)$ has
 no infinite paths then every supported model of~$T$ is stable.
\end{theorem}

Thus cycles and other infinite paths in $G^{pnn}(T)$ containing edges
that are not included in $G^{sp}(T)$ are harmless---they do not destroy the
match between stable models and supported models.  For instance, let~$T$ be
the pair of formulas
\beq\ba l
p\rar q,\\
((q\rar r)\rar r)\rar p.
\ea\eeq{p2}
The graph $G^{sp}(T)$ has two edges, $(q,p)$ and $(p,r)$, and
it is acyclic.  Consequently the stable models of~$T$ are identical to
its supported models~$\emptyset$, $\{p,q\}$.
The graph $G^{pnn}(T)$ is not acyclic in this case
because of the additional edge $(p,q)$.


\subsection{Loops} \label{ssec:l}

For any formula~$F$ and any set~$Y$ of atoms occurring in~$F$,
the ``negated external support'' formula $\hbox{NES}_F(Y)$ is defined
recursively, as follows:
\begin{itemize}
\item  for an atom~$A$, $\hbox{NES}_A(Y)$ is~$\bot$ if $A\in Y$, and $A$
      otherwise;
\item  $\hbox{NES}_\bot(Y) = \bot$;
\item  $\hbox{NES}_{F\land G}(Y) = \hbox{NES}_F(Y) \land \hbox{NES}_G(Y)$;
\item  $\hbox{NES}_{F\lor G}(Y) = \hbox{NES}_F(Y) \lor \hbox{NES}_G(Y)$;
\item  $\hbox{NES}_{F\rar G}(Y) = (\hbox{NES}_F(Y) \rar \hbox{NES}_G(Y))
                               \land (F\rar G)$
\end{itemize}
\cite[Section~3]{fer06}.
A set~$I$ of atoms occurring in~$F$ is a stable model of~$F$ iff it
satisfies both~$F$ and 
the \emph{loop formulas}
           \beq
              \bigwedge_{A\in Y}(A \rar \neg\hbox{NES}_F(Y))
           \eeq{dlf-gen}
           for all sets~$Y$ of atoms occurring in~$F$
           \cite[Theorem~2]{fer06}.  Furthermore, according to the same
theorem, there is no need to check all loop formulas~(\ref{dlf-gen}).  A
set~$Y$ of atoms occurring in~$F$ is called a \emph{loop} for~$F$ if the
subgraph of $G^{pnn}(F)$ induced by~$Y$ is strongly connected.  If~$I$
satisfies both~$F$ and the loop formulas~(\ref{dlf-gen}) for all loops~$Y$
of~$F$ then~$I$ is a stable model of~$F$.

The discussion in Section~\ref{ssec:sup} above suggests the question:
will the last result remain true if we replace the
graph $G^{pnn}(F)$ in the definition of a loop by the smaller
graph $G^{sp}(F)$?  The answer to this question is no.
A counterexample is given by the formula
\beq
(p\rar q)\land(((q\rar p)\rar p)\rar p)
\eeq{p3}
as~$F$, and $\{p,q\}$ as~$I$.  Indeed, the edges of the graph $G^{sp}(F)$
in this case are $(q,p)$ and $(p,p)$, and the sets~$Y$ for which the
subgraph of $G^{sp}(F)$ induced by~$Y$ is strongly connected are~$\{p\}$
and~$\{q\}$.  Calculations show that each of the formulas
$$\hbox{NES}_F(\{p\}),\ \hbox{NES}_F(\{q\})$$
is equivalent to
$\neg p\land\neg q$, so that each of the loop formulas
$$p\rar\neg \hbox{NES}_F(\{p\}),\ q\rar\neg \hbox{NES}_F(\{q\})$$
is a tautology.  Thus~$I$ is a model of~$F$ that satisfies these loop formulas,
although it is not stable.

The graph $G^{pnn}(F)$, on the other hand, has one more edge, $(p,q)$.
The subgraph of this graph induced by $\{p,q\}$ is strongly connected, and
the corresponding loop formula eliminates the model~$I$.

\subsection{Pointwise Stable Models} \label{ssec:psm}

Recall that a model~$I$ of a propositional theory~$T$ is stable if and only if
no proper subset of~$I$ satisfies the reduct~$F^I$ (Section~\ref{sec:ssc}).
We say that a model~$I$ of~$T$ is
\emph{pointwise stable} if there is no atom~$A$ in~$I$ such that
$I \setminus\{ A \}$ satisfies the reduct~$T^I$.
For example, $\{p,q\}$ is a pointwise stable model of $p\lrar q$.
Indeed, the reduct of $p\lrar q$ with respect to $\{p,q\}$ is $p\lrar q$;
it is not satisfied by any of the two sets obtained from $\{p,q\}$
by removing a single atom.

From published work on pointwise stable models \cite[Theorem~13]{fer09}
we can conclude that for any
finite propositional theory~$T$ such that the graph
$G^{pnn}(T)$ is acyclic, every pointwise stable model of~$T$ is stable.
The following theorem shows that the graph $G^{pnn}(T)$ in this statement
can be replaced by the smaller graph $G^{sp}(T)$:

\begin{theorem}\label{th2}
 For any propositional theory~$T$, if the graph $G^{sp}(T)$ has no
 infinite paths then all pointwise stable models of~$T$ are stable.
\end{theorem}

The additional generality of this theorem related to the use of
$G^{sp}(T)$ instead of $G^{pnn}(T)$ can be illustrated by formulas~(\ref{p2}).
Theorem~\ref{th2} shows that all pointwise stable models of that theory
are stable.


\subsection{Splitting} \label{ssec:s}

Splitting a logic program \cite{lif94e} allows us to relate its stable
models to stable models of its parts.  The form of splitting
described below is a special case of published results on splitting
first-order formulas~\cite{fer09a} and infinitary propositional
theories~\cite{har16a}, expressed in a form convenient for our present
purposes.

Let $\{P,Q\}$ be a partition of the set of atoms occurring in a
formula $F\land G$.  If
\begin{itemize}
\item[(i)]
 every atom that has a strictly positive occurrence in~$F$ belongs to~$P$, and
\item[(ii)]
 every atom that has a strictly positive occurrence in~$G$ belongs to~$Q$, and
\item[(iii)]
 every strongly connected component of $G^{pnn}(F\land G)$ is contained
 in~$P$ or in~$Q$,
\end{itemize}
then any set of atoms is a stable model of $F\land G$ if and only if
it is a stable model of each of the formulas
$$\ba c
F\land\bigwedge_{A\in Q}(A\lor\neg A),\ G\land\bigwedge_{A\in P}(A\lor\neg A).
\ea$$

This assertion will become incorrect, however, if we replace
$G^{pnn}(F\land G)$ in condition~(iii) by $G^{sp}(F\land G)$.  A
counterexample is given by formula~(\ref{p3}) as $F\land G$, $\{q\}$ as~$P$,
and $\{p\}$ as~$Q$.  Indeed, $\{p,q\}$ is a stable model of each of the
formulas
$$
\ba l
(p\rar q)\land(p\lor\neg p),\\
(((q\rar p)\rar p)\rar p)\land(q\lor\neg q),
\ea
$$
but not a stable model of~(\ref{p3}).

\section{Proofs of Theorems} \label{sec:proofs}

It is convenient to prove Theorem~\ref{th2} first.

For any formula~$F$, $\SPos(F)$ stands for the set of atoms
that have at least one strictly positive occurrence in~$F$.
For any propositional theory~$T$, $\SPos(T)$ is the union of the sets
$\SPos(F)$ over all formulas~$F$ in~$T$.

\begin{lemma}\label{lem:locally.tight.SPos.F} \emph{\cite[Electronic
    Appendix~C,
   Lemma~F]{lif13a}}
If an interpretation~$I$ satisfies a formula~$F$ then every
interpretation~$J$ such that
$\SPos(F^I) \subseteq J$ satisfies~$F^I$.
\end{lemma}

\begin{lemma}\label{lem:aux1}
 Let~$F$ be a propositional formula, let~$I,J$ be interpretations such
 that~$J \subset I$, and let $M$ be an atom in~$I\setminus J$
 such that
\begin{gather}
   \text{for every edge~$(M,A)$ of~$G^{sp}(F^I)$, $A\in J$.}
   \label{eq:1:lem:aux1}
\end{gather}
If~$M$ belongs to $\SPos(F^I)$ and~$J$ satisfies~$F^I$
then~$I \setminus \{ M \}$ satisfies~$F^I$ as well.
\end{lemma}

\medskip\noindent{\bf Proof.}
 Note first that, under the assumptions of the lemma,
 $I$ satisfies~$F$.  Indeed,
 otherwise $F^I$ would be~$\bot$, which contradicts the assumption that~$J$
 satisfies~$F^I$.

 The proof is by structural induction.
 Formula~$F$ is neither an atom nor~$\bot$.  Indeed, otherwise $F^I$
 would be an atom or~$\bot$ too; since $M\in\SPos(F^I)$, $F^I=M$.
 Since $M\in I\setminus J$, this contradicts the assumption that $J$
 satisfies~$F^I$.

 Let $F$ be $F_1\land F_2$, so
 that $F^I$ is $F_1^I\land F_2^I$.  Since $J$ satisfies~$F^I$,
 $J$ satisfies~$F^I_i$ \hbox{$(i=1,2)$}.  We need to show that
 $I\setminus\{M\}$ satisfies~$F^I_i$ as well.
 \emph{Case~1:} ${M \in \SPos(F_i^I)}$.  Since every rule
 of $F_i^I$ is a rule of $F^I$, $G^{sp}(F_i^I)$ is a
subgraph of~$G^{sp}(F^I)$; from~(\ref{eq:1:lem:aux1}) we can conclude that
$$\text{for every edge~$(M,A)$ of~$G^{sp}(F_i^I)$, $A\in J$.}$$
Then $I \setminus\{M\}$ satisfies~$F^I_i$ by the induction hypothesis.
\emph{Case~2:} ${M \not\in \SPos(F_i^I)}$.  Since $\SPos(F_i^I)$ is a subset
of~$I$, it follows that $\SPos(F_i^I) \subseteq I \setminus \{ M \}$.
On the other hand, $I$ satisfies $F_i$, because $J$ satisfies $F_i^I$. By
Lemma~\ref{lem:locally.tight.SPos.F}, these two facts imply that
$I \setminus \{ M \}$ satisfies $F_i^I$.

If $F$ is $F_1\lor F_2$ then the proof is similar.

Let $F$ be $F_1\to F_2$.  Then~$F^I$ is $F_1^I \to F_2^I$ and
${\SPos(F^I) = \SPos(F_2^I)}$, so that~${M \in \SPos(F_2^I)}$.
It follows that for every atom~$A$ in $\SPos(F_1^I)$,
the graph $G^{sp}(F^I)$ has an edge from~$M$ to~$A$.  Hence, by
assumption~\eqref{eq:1:lem:aux1}, every such atom~$A$ belongs to~$J$.  Thus
\beq
\SPos(F_1^I) \subseteq J.
\eeq{claim}
\emph{Case~1:} $J$ satisfies $F_2^I$.  Since every rule
 of $F_2^I$ is a rule of $F^I$, $G^{sp}(F_2^I)$ is a
subgraph of~$G^{sp}(F^I)$; from~(\ref{eq:1:lem:aux1}) we can conclude that
$$\text{for every edge~$(M,A)$ of~$G^{sp}(F_2^I)$, $A\in J$.}$$
By the induction hypothesis, it follows that
$I\setminus\{M\}$ satisfies~$F^I_2$,
and consequently satisfies~$F^I$.
\emph{Case~2:} $J$ does not satisfy $F_2^I$.
Then~$I$ does not satisfy~$F_1$.  Indeed, otherwise we would be able to
conclude by~(\ref{claim}) and
Lemma~\ref{lem:locally.tight.SPos.F} that~$J$ satisfies~$F_1^I$, which
contradicts the assumption that~$J$ satisfies~$F^I$.
Hence~${F_1^I = \bot}$, and $F^I$ is a tautology.

\medskip\noindent{\bf Proof of Theorem~\ref{th2}.}
Let~$I$ be a model of~$T$.  Assume that~$J$ is a proper
subset of~$I$ that satisfies~$T^I$; we need to show that
a subset satisfying~$T^I$ can be obtained from~$I$ by removing a single atom.

We will show first that
the set $I\setminus J$ contains an atom~$M$ satisfying
 condition~(\ref{eq:1:lem:aux1}).
\emph{Case~1:} $I\setminus J$ contains
an atom that is not a vertex of~$G^{sp}(T^I)$.  Then
condition~(\ref{eq:1:lem:aux1}) holds for that atom trivially.
\emph{Case~2:} all atoms in $I\setminus J$ are vertices of~$G^{sp}(T^I)$.
Assume that condition~(\ref{eq:1:lem:aux1}) is not satisfied for any of the
vertices~$M$ in $I\setminus J$, so that
\begin{center}
for every vertex~$M$ in~$I\setminus J$, $G^{sp}(T^I)$ has an edge
to some vertex~$A$ in~$I\setminus J$.
\end{center}
Since the set $I\setminus J$ is non-empty, it follows that the graph
$G^{sp}(T^I)$ has an infinite path.  But this is impossible, because
$G^{sp}(T^I)$ is a subgraph of $G^{sp}(T)$.

Take an atom~$M$ in $I\setminus J$ that satisfies
condition~(\ref{eq:1:lem:aux1}), and any formula~$F$ from~$T$.
If $M \in \SPos(F^I)$ then we conclude that
$I\setminus\{M\}$ satisfies~$F^I$
by Lemma~\ref{lem:aux1}.  Otherwise, $\SPos(F^I)\subseteq I\setminus\{M\}$,
and $I\setminus\{M\}$ satisfies~$F^I$ by Lemma~\ref{lem:locally.tight.SPos.F}.

\medskip\noindent{\bf Proof of Theorem~\ref{th1}.}
Let~$I$ be a supported model of a set~$T$ of nondisjunctive rules such
that the graph $G^{sp}(T)$ has no infinite paths; we need to show
that~$I$ is stable.  According to
Theorem~\ref{th2}, it is sufficient to check that~$I$ is pointwise stable.

Take any atom~$A$ in~$I$; we need to show that $I\setminus\{A\}$ is not
a model of~$T^I$.  Since~$I$ is supported,~$T$ contains a
nondisjunctive rule $\emph{Body}\rar A$ such that~$I$ satisfies
\emph{Body}.  The atom~$A$ has no strictly positive occurrences
in \emph{Body}; otherwise, $A,A,\dots$ would be an infinite path
in $G^{sp}(T)$.  Consequently
$$\SPos(\emph{Body}^I) \subseteq{\SPos(\emph{Body})
   \subseteq I \setminus \{ A \}}.$$
By Lemma~\ref{lem:locally.tight.SPos.F}, it follows that
$I \setminus \{ A \}$ satisfies~$\emph{Body}^I$.
Therefore $I \setminus\{ A \}$ does not satisfy the formula
$\emph{Body}^I \to A$, which belongs to~$T^I$.

\section{Conclusion}

The earliest use of positive dependency graphs for propositional
formulas~\cite{fer06} was related to the study of loops, and introducing
the~$G^{pnn}$ construction in that context rather than~$G^{sp}$ was
fully justified, as we saw in
Section~\ref{ssec:l}.  Using~$G^{pnn}$ in the theory of splitting was
justified as well (Section~\ref{ssec:s}).  Theorems~\ref{th1} and~\ref{th2}
show, on the other hand, that~$G^{sp}$ would be a better
tool for research on completion and on pointwise stable models.

The definitions of $G^{sp}$ and $G^{pnn}$,
as well as Theorems~\ref{th1} and~\ref{th2} and their proofs, can be
extended to infinitary propositional formulas.

The positive predicate dependency graph of a first-order formula can be
defined in two different ways as well, using either the ``sp'' approach
or the ``pnn'' approach.  The dependency graph defined by Bartholomew
and Lee \cite{bar19} is the sp-style predicate dependency graph for
first-order formulas with intensional functions.  Theorem~\ref{th1} above
is similar to
their Theorem~4.  It is less general in some ways (no variables and
quantifiers, no intensional functions) and more general in other ways
(the theory can be infinite and is not required to be in Clark normal form).

\section*{Acknowledgements}

Thanks to Paolo Ferraris, Joohyung Lee, Yuliya Lierler and the anonymous referees for comments on earlier versions of
this note.

\section*{Competing interests}
The authors declare none.

\bibliographystyle{acmtrans}
\bibliography{bib}
\end{document}